\documentclass{article}
\PassOptionsToPackage{numbers, compress}{natbib}
\usepackage[final]{neurips_data_2024}
\usepackage{makecell}
\usepackage{graphicx}
\usepackage{xcolor}
\usepackage{algorithm2e, setspace}
\SetKw{Continue}{continue}
\SetKw{Return}{return}
\RestyleAlgo{ruled}
\usepackage{amsmath}
\usepackage{array}
\newcolumntype{P}[1]{>{\centering\arraybackslash}p{#1}}
\newcolumntype{L}[1]{>{\flushleft\arraybackslash}p{#1}}
\usepackage{algpseudocode}
\usepackage{tabularx}

\usepackage{xcolor}
\graphicspath{ {./figures/} }
\usepackage{subcaption}

\usepackage[utf8]{inputenc}
\usepackage[T1]{fontenc}
\usepackage{hyperref}
\usepackage{url}
\usepackage{booktabs}
\usepackage{amsfonts}
\usepackage{nicefrac}
\usepackage{microtype}
\usepackage{xcolor}
\newcommand{\MOTIVE}{MOTI\texorpdfstring{$\mathcal{V}\mathcal{E}$}{VE}}

\title{\MOTIVE{}: A Drug-Target Interaction Graph For Inductive Link Prediction}

\author{%
   John Arevalo\thanks{Equal contribution} \quad Ellen Su$^*$ \quad Anne E. Carpenter \quad Shantanu Singh \\
   Broad Institute of MIT and Harvard \\
   \texttt{\{jarevalo, suellen, anne, shantanu\}@broadinstitute.org}
}

\begin{document}

\maketitle

\begin{abstract}\label{abstract}

Drug-target interaction (DTI) prediction is crucial for identifying new
therapeutics and detecting mechanisms of action. While structure-based methods
accurately model physical interactions between a drug and its protein target,
cell-based assays such as Cell Painting can better capture complex DTI
interactions. This paper introduces \MOTIVE{}, a \textbf{M}orphological
c\textbf{O}mpound \textbf{T}arget \textbf{I}nteraction \textbf{Graph} dataset
comprising Cell Painting features for $11,000$ genes and $3,600$ compounds,
along with their relationships extracted from seven publicly available
databases. We provide random, cold-source (new drugs), and cold-target (new
genes) data splits to enable rigorous evaluation under realistic use cases. Our
benchmark results show that graph neural networks that use Cell Painting
features consistently outperform those that learn from graph structure alone,
feature-based models, and topological heuristics. \MOTIVE{}
accelerates both graph ML research and drug discovery by promoting the
development of more reliable DTI prediction models. \MOTIVE{} resources are
available at \url{https://github.com/carpenter-singh-lab/motive}.

\end{abstract}

\section{Introduction}

High-quality graph benchmarking datasets propel graph machine learning (ML)
research. Providing a diversity of domains, tasks, and evaluation methods, they
allow for rigorous and extensive explorations of structured learning methods.
Still, gaps remain in the areas of scalability, network sparsity,
and generalizability under realistic data splits \citep{hu2021open}. These
challenges are particularly relevant in the biological domain. The
representation of the rich heterogeneity between entities\textemdash{}compounds,
genes, proteins, diseases, phenotypes, side effects, and more\textemdash{}is a
nontrivial task due to their varied units and terminology and highly
complex relational structure. This makes biological data an apt, challenging,
and bettering application for graph ML.

Next, effectively predicting drug-target interactions (DTIs), the relationships
between chemical compounds and their protein targets, remains a pressing
research area due to its relevance to drug discovery, drug repurposing,
understanding side effects, and virtual screening. The DTI task is challenging
due to the shortage of clean perturbational data and nonspecificity of these
interactions. Even as structure-based methods such as AlphaFold3 are
increasingly accomplished at making DTI predictions, they are mainly based on
molecular characteristics \citep{abramson2024accurate, zhou2024fragsite2}.
Experimental data uniquely captures complex biological interactions; the
morphological profiles, feature vectors that capture a cell's appearance, from
the Cell Painting (CP) assay have been shown to model the mechanism of action,
toxicity, and additional properties of compounds \citep{herman2023leveraging,
hamilton2017inductive}.

To address the challenges of graph ML, biological data representation, and DTI,
we introduce a publicly available dataset, \MOTIVE{}, which enhances a graph of
compound and gene relations with features from the JUMP Cell Painting dataset
\citep{chandrasekaran2023jump}. As there is currently no compound-gene graph
dataset containing empirical node features, \MOTIVE{} will be extremely useful
for inductive graph learning \citep{hamilton2017inductive} (generalizing to
newly connected nodes), cold start recommendations \citep{schein2002methods}
(generalizing to isolated nodes), and zero-shot scenarios
\citep{qin2020generative} (generalizing to isolated node pairs). In many
domains, making predictions for least-known entities are the most useful
real-world applications \citep{ezzat2018computational}. Thus, we accompany
\MOTIVE{} with a rigorous framework of data splitting, loading, and evaluation.
This work advances both DTI by incorporating a new modality of information and
the strength of graph ML, as it rises to the challenge of knowledge
generalization for inductive link prediction.

\section{Related work}

Although many graph-based datasets exist, \citet{hu2021open} notes that there is
a trade-off between scale and availability of node features. The Open Graph
Benchmark Library (OGBL) thus contributed \verb|ogbl-ppa|, \verb|ogbl-collab|,
and \verb|ogbl-citation2|, all large-scale, feature-based link prediction
datasets. The node features in each of these datasets are 58- or 128-dimensional
and are a one-hot representation of the protein type in \verb|ogbl-ppa| or
\verb|Word2Vec|-based representations of an author's publications or of a
paper's contents in \verb|ogbl-collab| and \verb|ogbl-citation2| respectively.
The benchmarking results showed a continued reliance on model learning from
previous connections rather than features, as evidenced by the high performance
of \verb|Node2Vec| embeddings in OGBL tasks, and indicated a need for richer
features. The authors also reported that the graph neural network (GNN) models
underperformed in the link prediction task when using mini-batch training rather
than whole batch and called for improvement in this area for future scalability
when learning on large datasets. In addition, the latter two datasets split
their data by time metadata associated with each edge, and did not explore cold
start splits. Evidently, the field still requires graph datasets that 1) are
large-scale, 2) include information-rich features, 3) are accompanied by
graph-based data splits (not metadata-based), and 4) are flexibly trained with
mini-batch sampling. We prioritized all four goals during the curation of
\MOTIVE{} and in our experimental design.

As Knowledge Graphs (KGs) have emerged as powerful tools for representing and
learning from network-based data \citep{caufield2023building}, they have often
been applied to biological and chemical tasks such as drug discovery, structure
prediction, and DTI \citep{chandak2023building, himmelstein2017systematic,
huang2021therapeutics}. Recent surveys on graph-based methods for DTI prediction
\citep{zhang2023survey, zhang2022graph} show few efforts represent drugs and
genes with external features. Two related approaches are
\citet{schwarz2024ddos}, which used gene expression as input to the last layer
in a late-fusion manner, and \citet{balamuralidhar2023prediction}, which used
topological features as node representations. While
\citet{balamuralidhar2023prediction} does enable inductive predictions on newly
connected nodes, it fails to make inferences on completely isolated nodes in
cold start scenarios. To our knowledge, no graph-based dataset exists where the
features of compounds and gene nodes are represented by their morphological
profiles.

Meanwhile, in biological image analysis, fluorescence labeling of cells now
allows for the visualization of cell morphology, internal structures, and
processes at unprecedented spatial and temporal resolution. Imaging precisely
captures the changes to a cell after it has undergone a chemical or genetic
perturbation and sheds light on relationships such as DTI, functions, and
mechanisms. Additionally, the recent publication of the JUMP Cell Painting
dataset \citep{chandrasekaran2023jump} now provides $136,000$ chemical and
genetic perturbational profiles for the DTI task. In recent applications of CP
to DTI, \citet{rohban2022virtual} matched compounds to a small set of genes
based on morphological feature vector similarities, and we will extend this work
by using machine learning to capture non-linear compound-gene relationships for
a much larger gene set. Next, \citet{herman2023leveraging} developed a deep
learning model to predict toxicity assays from chemical structures and
morphological profiles but did not exploit the connectivity network of compounds
and gene interactions. Our method builds on this approach by incorporating the
morphological profiles of both compounds and genes in a graph and using the
message-passing framework to leverage such network connectivity. Recently,
\citet{iyer2024cell} formulated the DTI task as a binary classification of
gene-compound pairs under different data splits and developed a
transformer-based learning approach to predict drug targets from Cell Painting
profiles. Although this approach is similar to our proposed setup, their dataset
contains fewer nodes, 302 compounds and 160 genes, and does not include
gene-gene and compound-compound interactions.

The graph ML community needs large-scale datasets with rich, empirical node
features, and well-defined data splitting, loading, and evaluation procedures
that advance scalability (training in batches) and generalizability (inductive
link prediction). The drug discovery community is motivated by reliable DTI
predictions, especially for newly discovered or assayed compounds and genes with
no known relationships. The biological image analysis community is likewise
curious about the reach, applicability, and predictive power of the
morphological profiles of perturbed cells. To address these needs across
multiple domains, we contribute the \MOTIVE{} dataset.

\section{The \MOTIVE{} dataset}\label{sec:dataset}

The \MOTIVE{} dataset leverages drug-target knowledge graphs and image-based
profiles of chemically or genetically perturbed cells. It comprises $11,509$
genes, $3,632$ compounds, and $303,156$ compound-compound, compound-gene, and
gene-gene interactions collected from seven publicly available databases.

\subsection{Morphological profiles extraction}

Every gene (each of which produces a particular protein) and compound is
represented with its image-based profile from the JUMP CP dataset
\citep{chandrasekaran2023jump}. JUMP CP images capture a cell's morphology after
being perturbed by a chemical compound or a genetic edit. Each type of genetic
perturbation, CRISPR knockouts or ORF over-expressions, involves a different set
of genes. We created and published a version of \MOTIVE{} for each gene set but
chose to use the ORF genes for the analysis presented in this paper due to
slightly stronger downstream performance. The morphological profiles were
extracted from each image in the JUMP CP dataset using CellProfiler
\citep{stirling2021cellprofiler}, a software that segments individual cells in
the image and measures thousands of features for each cell. Then, the data was
prepared according to the protocols in \citet{arevalo2024evaluating} and
\citet{chandrasekaran2023jump}, which extensively optimize separate pipelines
for compound and gene perturbations.

For compound perturbations, we first filtered out features with low variance,
then applied median absolute deviation normalization, a rank-inverse normal
transformation (INT), and Harmony \citep{Korsunsky2019FastSA} to reduce the
batch effects. We then selected final features based on correlation analysis.
For genetic perturbations, we subtracted the mean vector per well from each
feature vector to account for well position effects, replaced the INT
transformation with an outlier removal step, and aggregated the replicates
(usually five) of each perturbation using median profiles to make the
representations robust to low-quality images. In our experience, these comprised
less than $5\%$ of the data and were uncorrelated across replicates on different
plates. After the correction and preprocessing steps, each compound and gene is
represented by a 737-dimensional and 722-dimensional vector, respectively. The
processing pipelines for the morphological profiles are available at
\url{https://github.com/broadinstitute/jump-profiling-recipe/tree/v0.1.0}.

\subsection{Annotation collection} \label{sec:datacollection}
\begin{table}[ht]
    \centering
\begin{tabular}{lllll}
\toprule
Database  & \vtop{\hbox{\strut compound}\hbox{\strut gene}} & \vtop{\hbox{\strut compound}\hbox{\strut compound}} & \vtop{\hbox{\strut gene}\hbox{\strut gene}} & \vtop{\hbox{\strut compound}\hbox{\strut identifier}} \\
\midrule
BioKG \citep{walsh2020biokg} & \checkmark & \checkmark & \checkmark & DrugBank \\
DGIdb \citep{freshour2020integration} & \checkmark &  &  & ChEMBL \\
DRHub \citep{corsello2017drug} & \checkmark &  &  & PubChem \\
Hetionet \citep{himmelstein2017systematic} & \checkmark & \checkmark & \checkmark & DrugBank \\
OpenBioLink \citep{breit2020openbiolink} & \checkmark &  & \checkmark & PubChem \\
PharMeBINet \citep{knigs2022heterogeneous} & \checkmark & \checkmark & \checkmark & DrugBank \\
PrimeKG \citep{chandak2023building} & \checkmark & \checkmark & \checkmark & DrugBank \\
\bottomrule
\end{tabular}
    \caption{Databases integrated in \MOTIVE{}. We extracted all compound-gene,
    compound-compound and gene-gene annotations. Compound IDs were mapped to the
    InChIKey representations, and gene IDs were mapped to the NCBI gene
    symbols.}
    \label{tab:annotation_db_list}
\end{table}

We aggregated the compound-compound, compound-gene and gene-gene annotations in
\MOTIVE{} from seven publicly available databases listed in Table
\ref{tab:annotation_db_list}. Our choice followed the comprehensive review of
\citet{bonner2022review}, which categorizes relevant biomedical datasets and KGs
based on the entities they contain and how well they fit to specific tasks. We
unified the compound IDs by mapping them to their core molecular skeleton
identifiers (InChIKey) using the MyChem \citep{lelong2022biothings} and UniChem
\citep{chambers2013unichem} databases and unified the gene IDs using their NCBI
gene symbols \footnote{\url{https://www.ncbi.nlm.nih.gov/gene/}}. We then
combined all of the pairwise interactions we collected across databases with the
JUMP CP features, keeping only the pairs where both entities had existing
features and using only the pairs for which we had interactions.

\subsection{Graph construction} \label{sec:graphconstruction}

Since DTI is defined by identifying the protein targets for a given drug, we
assigned nodes associated with compounds as source nodes and nodes associated
with genes as target nodes. We refer to compound-gene, compound-compound, and
gene-gene as source-target, source-source, and target-target interactions
respectively. We defined each source or target as a node and each interaction as
an undirected edge. The representation of each node was set as its processed
morphological feature vector as described in Section \ref{sec:datacollection}.

We defined our graph $G = (\mathcal{V}, \mathcal{E})$ as follows. $\mathcal{V} =
\mathcal{S} \cup \mathcal{T}$ is the union of the sets of sources and targets in
our dataset. Every $s \in \mathcal{S}$ and $t \in \mathcal{T}$ is represented by
a feature vector $x_s \in \mathbb{R}^n$ and $x_t \in \mathbb{R}^m$,
respectively. The edge set $\mathcal{E} = SS \cup ST \cup TT$ is the union of
our source-source, source-target, and target-target edges, where each edge is
defined as a pair of node indices: $(s_u, s_v) \in SS$, $(s_u, t_v) \in ST$, and
$(t_u, t_v) \in TT$. We defined four different graph structures to assess the
information added by each of our edge types: a \emph{bipartite} graph which only
includes $ST$ edges, a source-expanded (\emph{s\_expanded}) graph which includes
$ST$ and $SS$ edges, a target-expanded (\emph{t\_expanded}) graph which includes
$ST$ and $TT$ edges, and a source and target expanded (\emph{st\_expanded})
graph which includes all three edge types (statistics in Table
\ref{tab:graphdegree}). Note that, unless otherwise stated, our evaluated models
received the \emph{st\_expanded} graph as it maximized the available information
in \MOTIVE{}.

\begin{table}[ht]
    \centering
\begin{tabular}{llllll}
\toprule
\makecell[l]{Graph\\ Type} & \# Nodes & \makecell[l]{\# ST\\ Edges} & \makecell[l]{\# Other\\ Edges} & \makecell[l]{Avg. Node\\ Degree} & \makecell[l]{Med. Node\\ Degree} \\
\noalign{\smallskip}\hline\noalign{\smallskip}
bipartite &
\begin{tabular}{l}S: 2,961 \\ T: 4,505\end{tabular}
& 24,798 & \begin{tabular}{l}SS: 0 \\ TT: 0\end{tabular} &
\begin{tabular}{l}S: 8.4 \\ T: 5.5\end{tabular} & \begin{tabular}{l}S: 3.0 \\ T: 3.0\end{tabular}\\

s\_expanded &
\begin{tabular}{l}S: 3,632 \\ T: 4,505\end{tabular}
& 24,798 & \begin{tabular}{l}SS: 75,330 \\ TT: 0\end{tabular} &
\begin{tabular}{l}S: 48.3 \\ T: 5.5\end{tabular} & \begin{tabular}{l}S: 19.5 \\ T: 3.0\end{tabular}\\

t\_expanded &
\begin{tabular}{l}S: 2,961 \\ T: 11,509\end{tabular}
& 24,798 & \begin{tabular}{l}SS: 0 \\ TT: 203,028\end{tabular} &
\begin{tabular}{l}S: 8.4 \\ T: 37.4\end{tabular} & \begin{tabular}{l}S: 3.0 \\ T: 25.0\end{tabular}\\

st\_expanded &
\begin{tabular}{l}S: 3,632 \\ T: 11,509\end{tabular}
& 24,798 & \begin{tabular}{l}SS: 75,330 \\ TT: 203,028\end{tabular} &
\begin{tabular}{l}S: 48.3 \\ T: 37.4\end{tabular} & \begin{tabular}{l}S: 19.5 \\ T: 25.0\end{tabular}\\

\bottomrule
\end{tabular}
    \caption{Statistics for the four graph structures.}
    \label{tab:graphdegree}
\end{table}

\subsection{Data splitting} \label{datasplitting}
\begin{figure}[ht]
\centering
\includegraphics[width=0.75\textwidth]{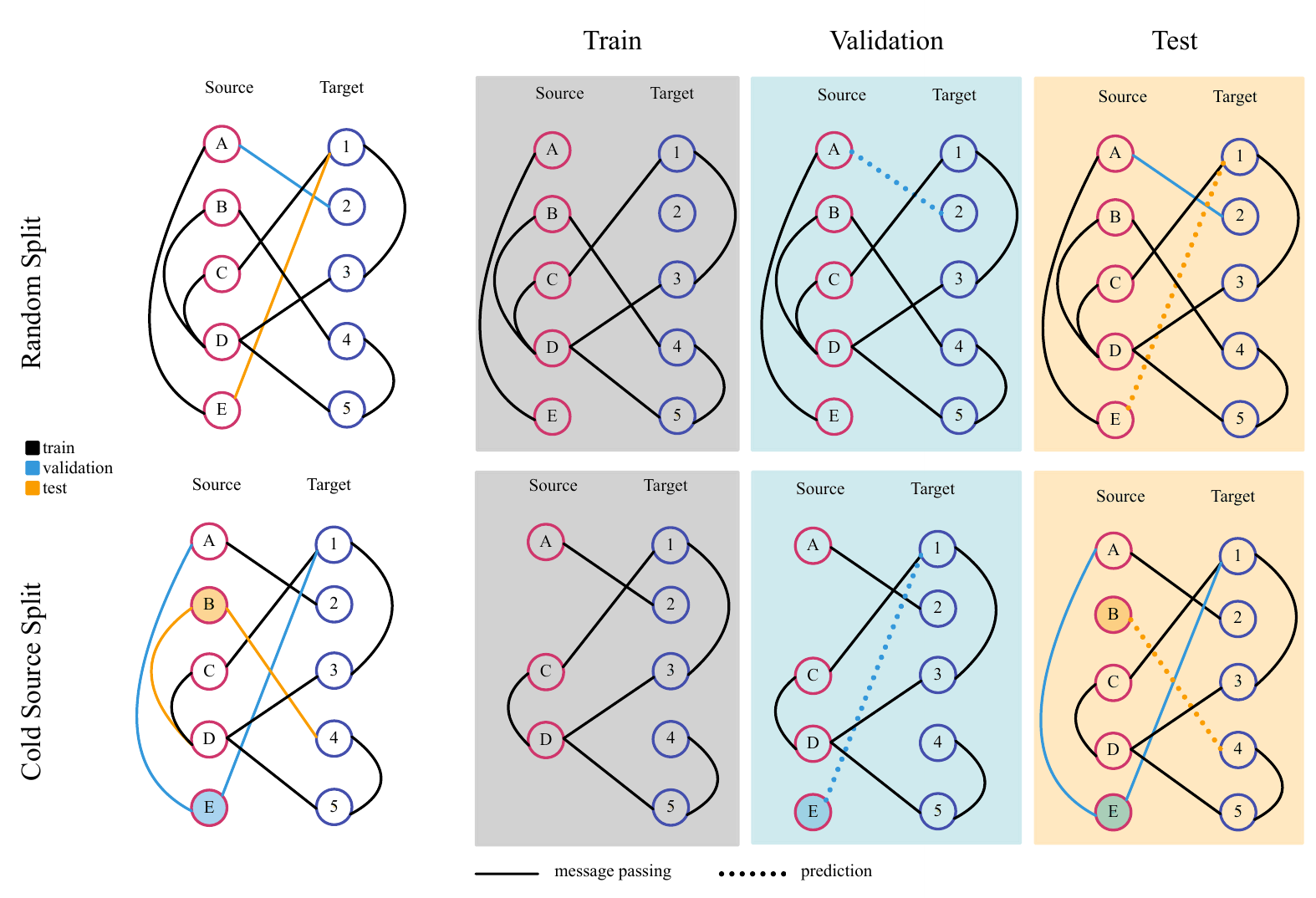}
\caption{Schematic of the random split (top row) and cold-source split (bottom
row). The left-most graph illustrates the actual partitioning of the edges, and
the three graphs to the right show which nodes and edges are visible to the GNN
models during training, validation, and testing. The number of edges in each
partition is not representative of our true 70/10/20 ratios. Cold-target split
is symmetrical to cold-source split. The model aggregates neighbor features via
the message-passing edges (solid lines) and makes predictions on the supervision
edges (dotted lines).}
\label{fig:datasplit}
\end{figure}

We developed two different heuristics to split our graph into training,
validation, and test sets based on the random split and cold start split
originally defined for recommendation systems \citep{schein2002methods}.

In the random split scenario, we used a 70/10/20 ratio to randomly select and
split every $ST$ edge into train, validation, and test sets and included the
full set of $SS$ and $TT$ edges in training (Figure \ref{fig:datasplit}). In
this case, the link prediction task is transductive, as the nodes are fixed from
the start of training and the model will predict edges on entities it has
learned on. Negative edges are sampled for each batch of data by randomly
selecting source-target pairs that are not in the $ST$ edges.

The cold start split allows for inductive link prediction, as it involves
predicting edges where at least one node was not present during training. We
applied the cold start split to either the source or target nodes, denoted by
cold-source split or cold-target split. In cold-source split (second row of
Figure \ref{fig:datasplit}), every source node in our graph is randomly labeled
as either train, validation, or test in a 70/10/20 ratio. All $ST$ edges are
subsequently labeled in accordance with the label of its source node. Next, all
$SS$ edges are labeled by their most conservative labels: any edge with at least
one test source is labeled test, any edge with no test sources but at least one
validation source is labeled validation, and any edge with two train sources is
labeled train. All $TT$ edges are included in training. This data split prevents
data leakage of any kind and enforces that at validation or test time, the model
has never seen any source node in the $ST$ edges. Additionally, negative edges
are sampled for each batch of data by our custom sampler class, built for
inductive learning, which only considers the sampled source nodes and all of
target nodes, in the $ST$ edges. Details about our negative sampling algorithm
can be found in Section \ref{sec:negSampling}.

Cold-target split is constructed in a symmetrical manner to cold-source split.
The cold splits are useful for considering how the model will generalize to
unseen and isolated sources and targets. When the node embeddings represent
real-world features, models trained under this stringency will be especially
useful in predicting novel yet relevant relationships from empirical data.

\subsection{Negative sampling algorithm} \label{sec:negSampling}
\begin{algorithm}[]

\caption{Negative sampling for cold-source split}\label{alg:negSampling}
\KwIn{\begin{tabular}[t]{l}
      $ST = \{(s_u,t_v) \mid \exists \text{ an edge between source } u \text{ and target }v\}$: all ground truth $ST$ edges\\
      $B_P \subseteq ST$: positive supervision edges in batch\\
      $r \in \mathbb{Z}_{>0}$: negative sampling ratio\\
      $f(A, k)$: sample $k$ elements from set $A$ \\
      $n$: number of sampling tries
      \end{tabular}}
\KwOut{$B'$: positive and negative supervision edges in batch}
$c \gets \lvert B_P\rvert$ \Comment{\# positives}\;
$S_B \gets \{s \mid \exists (s, t) \in B_P\}$\;
$T_{ST} \gets \{t \mid \exists (s, t) \in ST\}$\;

\For{$i:=0$ \KwTo $n$}{
    $B_N \gets f\left(\{(s, t) \mid s\in S_B \land t \in T_{ST}\}, 2cr\right) $\Comment{Sampled negative edges}\;
    $B_N \gets B_N \setminus ST$ \;
    \If {$\lvert B_N\rvert \geq cr$}{
        $B'_N \gets f\left(B_N, cr\right)$\;
        $B' \gets B_P \cup B'_N$\;
        \Return $B'$\;
    }
    }
\end{algorithm}

Scalable gradient-based methods process data in randomly sampled batches, or
subgraphs in the graph domain. Link prediction approaches commonly generate
negative edges by dynamically sampling disconnected nodes. To prevent data
leakage, the sampled negative edges during training must not include cold
instances from validation or testing. However, we observed that the negative
sampling strategy in PyG \citep{fey2019fast}, a widely used GNN framework, lacks
the granularity to control the sampled population. To address this limitation,
we developed a custom negative sampling algorithm that guarantees test node
isolation and enables proper evaluation of link prediction models.

We used a negative sampling ratio ($r$) of 1:1 during training and 1:10 during
testing. For random split, our algorithm samples the negative edges between all
unique source and target nodes within the $ST$ edges in the batch. For
cold-source split, the head of the negative edge is sampled from the unique
sources in the supervision $ST$ edges in the batch, and the tail of the negative
edge is sampled from $ST$ edges in the batch. Algorithm \ref{alg:negSampling}
details the cold-source split negative sampling procedure. The process is
symmetrical for leave-out-target.

\subsection{Models}

We experimented with three different types of GNN convolutional layers in our
model, each with a unique learning algorithm. The algorithms differ in how they
incorporate the features of neighbor nodes into a node's learned representation.
GraphSAGE (Graph SAmple and aggreGatE) \citep{hamilton2017inductive} achieves
inductive representation learning on large graphs by sampling from neighbor node
representations and applying an aggregation function (e.g. a weighted average)
to compute each node's hidden embedding. GIN (Graph Isomorphism Network)
\citep{xu2019powerfulgraphneuralnetworks} adds a Multilayer Perceptron (MLP) to
represent the composition of neighbor node features and achieves discriminative
power equal to the Weisfeiler Lehman graph isomorphism test. Finally, GATv2
(Graph Attention Network v2) \citep{brody2022attentivegraphattentionnetworks}
implements dynamic graph attention which weights neighbor representations
according to each query of interest. We constructed separate models using each
of these GNN algorithms (GraphSAGE$_{CP}$, GIN$_{CP}$, and GATv2$_{CP}$) as the
convolution layer, and we used the JUMP \underline{C}ell \underline{P}ainting
features as the nodal representations for each. We chose to only initialize the
node features using their image-based features (as opposed to chemical
structure-based or protein structure-based) so that compounds and genes would
share a modality of representation. This ensures similar statistical properties
between node types and makes it easier to discover cross-modal relationships
\cite{Srivastava2012MultimodalLW}.

All three GNN models share the same architecture. First, the input embeddings of
every source and target node are initialized and fixed as their respective
feature vectors $x_s$ and $x_t$. Two linear layers then transform the source and
target node embeddings into the same embedding space. These  transformed
embeddings pass through two GNN layers (chosen from one of the three graph
convolutional algorithms above), separated by a leaky ReLU activation function
for nonlinearity. Broadly, each GNN layer combines the feature vector of the
node of interest with some aggregation of the feature vectors from the neighbors
of the node of interest. The aggregation function changes according to the three
GNN algorithms. Isolated nodes rely solely on CP features due to the lack of
neighboring signals. The two GNN layers indicate that the feature vectors of
neighbor nodes within a radius of 2 will be used to compute the hidden embedding
for each node. An additional skip connection feeds the output of the first GNN
layer (an aggregation of the feature vectors in the neighborhood of radius 1)
into the final embedding, prioritizing shorter distance neighbors. Finally, a
classifier head outputs the sigmoid-transformed dot product of the embeddings
for each source and target node pair in the supervision edge set. Algorithm
\ref{alg:gnn} in Appendix \ref{CPalg} further describes the forward pass of our
GNNs to make the link predictions.

We benchmarked our model with a featureless graph-based model that randomly
initializes the source and target input embeddings. We opted to use the simple
GraphSAGE convolutional layer in our model because all of the node features are
random vectors in this case. This model, GraphSAGE$_{embs}$, shares the same
architecture as the other GNNs. Importantly, GraphSAGE$_{embs}$ does not have
informative feature vectors for each source and target node at the start of
training. We also applied one baseline heuristic for topology-based link
predictions, and two baseline models for feature-based link predictions. For the
former, we used the shortest path between source and target nodes as a proxy for
their similarity to evaluate the predictive power of graph structure alone. For
the latter, we implemented a Bilinear model and an MLP, which both rely solely
on node features to predict links. The Bilinear model learns a mapping $y=x^TWz$
to transform two sets of feature vectors into the same embedding space, then
computes the similarity between the vectors. The MLP learns hidden embeddings of
the feature vectors via two fully connected layers and ReLU activation functions
for nonlinearity, then uses a Bilinear head to make link predictions. These
latter models ignore the known relational information between entities when
predicting links. Across the seven models, we could readily evaluate the
performance enhancements coming from adding node features or graph structure and
determine the most useful learning approach for each of our scenarios. See
Appendix \ref{sec:experimental} for experimental details and evaluation metric
choices.

\section{Results}\label{results}

\subsection{DTI prediction improves with CP features}

\begin{table}[ht]
    \centering
    \begin{tabular} {P{2cm}P{2.4cm}P{2.3cm}P{2.3cm}P{2.3cm}}
    \toprule
    \textbf{Input} & \textbf{Model} & \textbf{F1} & \textbf{Hits@500}& \textbf{Precision@500}\\
    \hline
    & GIN$_{CP}$ & \textbf{0.5238{\color{lightgray} $\pm$ 0.040}} & \textbf{0.4552 {\color{lightgray}$\pm$0.017}} & \textbf{0.9920 {\color{lightgray} $\pm$0.005}} \\
    Graph+CP & GATv2$_{CP}$ & 0.4169{\color{lightgray} $\pm$0.007} & 0.2219{\color{lightgray} $\pm$0.018} & 0.8452{\color{lightgray} $\pm$0.045} \\
    & GraphSAGE$_{CP}$ & 0.3836{\color{lightgray} $\pm$0.029} & 0.2637{\color{lightgray} $\pm$0.023} & 0.9056{\color{lightgray} $\pm$0.012} \\

    \hline
    CP & MLP & 0.3829{\color{lightgray} $\pm$0.008} & 0.2545{\color{lightgray} $\pm$0.011} & 0.8456{\color{lightgray} $\pm$0.015} \\
    & Bilinear & 0.1703{\color{lightgray} $\pm$0.003} & 0.0213{\color{lightgray} $\pm$0.001} & 0.1812{\color{lightgray} $\pm$0.008} \\

    \hline
    Graph & GraphSAGE$_{embs}$ & 0.3456{\color{lightgray}$\pm$0.006} & 0.2254{\color{lightgray}$\pm$0.013} & 0.8476{\color{lightgray}$\pm$0.029} \\

    & Shortest Path & \textemdash &0.0025&0.0 \\
    \bottomrule
    \end{tabular}
    \caption{Test metrics (all with maximum value$=$1), averaged over 5 runs,
    for all models in the random split scenario. GNNs initialized with Cell
    Painting data are indicated with a $CP$ subscript. The shortest path
    heuristic is fixed across runs, and thus does not have standard deviation
    values. In addition, only rank based metrics were computed for shortest path
    since there is no classification threshold. The metric parameter k=500
    corresponds to the top $1\%$ of the test edges.}

\label{tab:transductive}
\end{table}

First, we evaluated all of our models on transductive link predictions (random
split). We used the shortest path heuristic, a non-learning baseline, to predict
links based on graph topology (last row of Table \ref{tab:transductive}). We
scored each positive and negative $ST$ edge in our test set by the shortest path
length between the two nodes and computed rank based metrics based on these
scores. This heuristic fails to effectively predict links on \MOTIVE{}, proving
the task nontrivial and indicating a need for learning-based methods. The low
prediction scores from the Bilinear and MLP models (rows 4 and 5 of Table
\ref{tab:transductive}) also indicate that predicting edges purely based on the
similarity of source and target node features is inadequate. Still, these models
significantly outperform the non-learning baseline, signaling the information
contained in the node features.

Next, we see a large score increase from the feature-based benchmarks to the GNN
models, which make use of the graph structure and CP node features. The
predictions made by the learned representations of source and target nodes from
topology alone (GraphSAGE$_{embs}$) achieve a Precision@500 score of 84.76\%,
which highlights the richness of the relational information between nodes.
Finally, the GIN$_{CP}$ model achieves the best performance across metrics,
obtaining a Hits@500 score of 45.54\% and a Precision@500 score of 99.20\%. This
result supports our hypothesis and demonstrates that adding CP node features to
graph structure benefits transductive link prediction.

\subsection{Inductive link prediction benefits from graph structure and CP features}

\begin{table}[ht]
    \centering

\begin{tabular}{P{1cm}P{2.2cm}P{2.8cm}P{2.8cm}P{2.8cm}}
    \toprule
    \textbf{Split}&\textbf{Model} & \textbf{F1} & \textbf{Hits@500} & \textbf{Precision@500}\\
    \midrule
       & GIN$_{CP}$ & \textbf{0.2827{\color{lightgray}$\pm$0.018}} & \textbf{0.1078{\color{lightgray}$\pm$0.008}} & \textbf{0.5872{\color{lightgray}$\pm$0.028}} \\
       & GraphSAGE$_{CP}$ & 0.2283{\color{lightgray}$\pm$0.014} & 0.0439{\color{lightgray}$\pm$0.013} & 0.3120{\color{lightgray}$\pm$0.079} \\
Source & GATv2$_{CP}$ & 0.2433{\color{lightgray}$\pm$0.009} & 0.0330{\color{lightgray}$\pm$0.002} & 0.2476{\color{lightgray}$\pm$0.004} \\
       & MLP & 0.1876{\color{lightgray}$\pm$0.006} & 0.0269{\color{lightgray}$\pm$0.009} & 0.2168{\color{lightgray}$\pm$0.063} \\
       & Bilinear & 0.1593{\color{lightgray}$\pm$0.004} & 0.0145{\color{lightgray}$\pm$0.003} & 0.1256{\color{lightgray}$\pm$0.029} \\
    \hline

       & GIN$_{CP}$ & \textbf{0.3916{\color{lightgray}$\pm$0.016}} & \textbf{0.2744{\color{lightgray}$\pm$0.020}} & \textbf{0.9624{\color{lightgray}$\pm$0.034}} \\
       & GraphSAGE$_{CP}$ & 0.3494{\color{lightgray}$\pm$0.012} & 0.1988{\color{lightgray}$\pm$0.013} & 0.7056{\color{lightgray}$\pm$0.031} \\

Target & GATv2$_{CP}$ & 0.2805{\color{lightgray}$\pm$0.007} & 0.1966{\color{lightgray}$\pm$0.019} & 0.9540{\color{lightgray}$\pm$0.043} \\
       & MLP & 0.3408{\color{lightgray}$\pm$0.014} & 0.1837{\color{lightgray}$\pm$0.010} & 0.6552{\color{lightgray}$\pm$0.025} \\
       & Bilinear & 0.1396{\color{lightgray}$\pm$0.001} & 0.0122{\color{lightgray}$\pm$0.001} & 0.1024{\color{lightgray}$\pm$0.004} \\
\bottomrule
\end{tabular}

\caption{Test metrics (all with maximum value$=$1), averaged over 5 runs, for
all feature-based models. The top five rows show the cold-source split results
and the bottom five rows show the cold-target split results.}
\label{tab:inductive}
\end{table}

We evaluated the models on inductive link prediction tasks using the cold data
splits, which require informative node features as the left out nodes are
completely isolated during evaluation \citep{hamilton2017inductive}. In this
scenario, GraphSAGE$_{embs}$ is not applicable as it learns representations for
nodes using projection layers and is thus limited to making predictions for
nodes that it has already seen during training (i.e. transductive link
predictions). The shortest path heuristic is also not applicable, as it is
unable to handle isolated nodes unreachable by any path.

During inductive link prediction, all of the models only have access to the
feature vectors of left-out nodes, as they are isolated and without edge
connections. For both cold-source and cold-target split, GIN$_{CP}$ greatly
outperformed all other models. The improved predictions by GIN$_{CP}$ can be
attributed to it having learned improved source and target node embeddings by
leveraging both the node features and the relational information present in the
graph structure during training. Except for the F1 score of the GATv2 model in
the cold-target split, all of the GNN models outperform the feature-based
models. This supports our hypothesis that knowing the relationships between
sources and targets improves DTI prediction.

Inductive link prediction is useful in real world applications as the model is
learning about and making predictions on entities that we previously knew very
little about. The cold-source split is especially relevant for drug discovery,
as it simulates the scenario where all genes are known and the DTI model is
tasked with identifying new compounds associated with certain disease-related
genes. Furthermore, empirical features, such as morphological profiles, prove to
be valuable node representations. If we collect experimental data for a newly
discovered compound or unexplored gene, GIN$_{CP}$ and similar models would be
able to leverage known relationships between other compounds and genes to make
better inductive link predictions for the new entity. This underscores the value
of integrating assay-based data into graph-based models to enhance their
predictive capabilities.

\subsection{Ablation studies with graph structure} \label{ablation}

\begin{figure}[ht]
    \centering
    \includegraphics[width=0.9\textwidth]{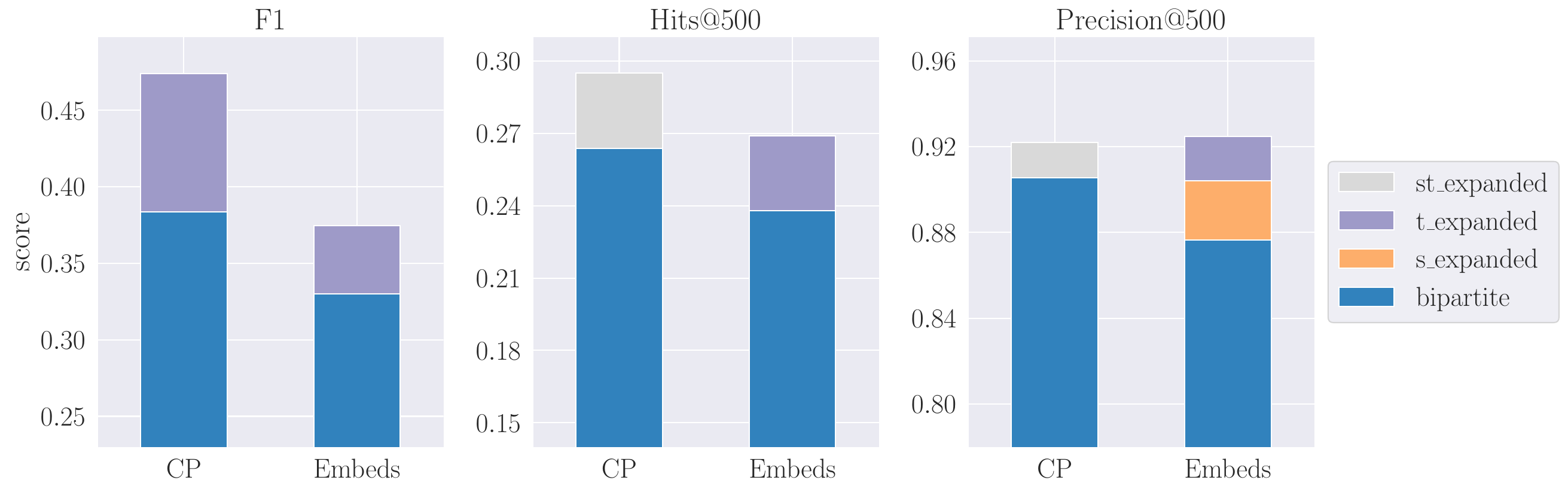}
    \caption{Test metrics for GraphSAGE$_{CP}$ and GraphSAGE$_{embs}$
        for all four graph structures, with random data splits, and averaged
        over 5 runs. The colored stratifications of each bar show the
        decreasing performances of the models as edge types are removed from
        the graph. Note that the bars of each color are overlaid onto each
        other in the order specified in the legend, such that a structure color
        will only appear if it obtained worse performance than the previous
        structure.}
    \label{fig:ablation}
\end{figure}

We investigated the contribution of each edge type (source-target,
source-source, and target-target) by comparing the performances of the GraphSAGE
models using the four graph structures (Figure \ref{fig:ablation}). While this
analysis can apply to any of the GNN models, we chose GraphSAGE such that we
could compare the feature-based and embeddings-based cases. We see that
GraphSAGE$_{CP}$ outperforms GraphSAGE$_{embs}$ for almost every graph structure
and metric. More analyses should be done to investigate how the difference in
performance fluctuates with the addition of edge types in order to understand
how the usefulness of the node features fluctuates with graph sparsity. From
these initial results, it appears that as the graph becomes denser with the
addition of more edge types, the structural information begins to dominate,
reducing the relative contribution of the node features. These findings
highlight the complementary nature of node features and graph structure in the
link prediction task.

\subsection{Zero-shot prediction potential}
\begin{figure}[ht]
    \centering
    \includegraphics[width=0.8\textwidth]{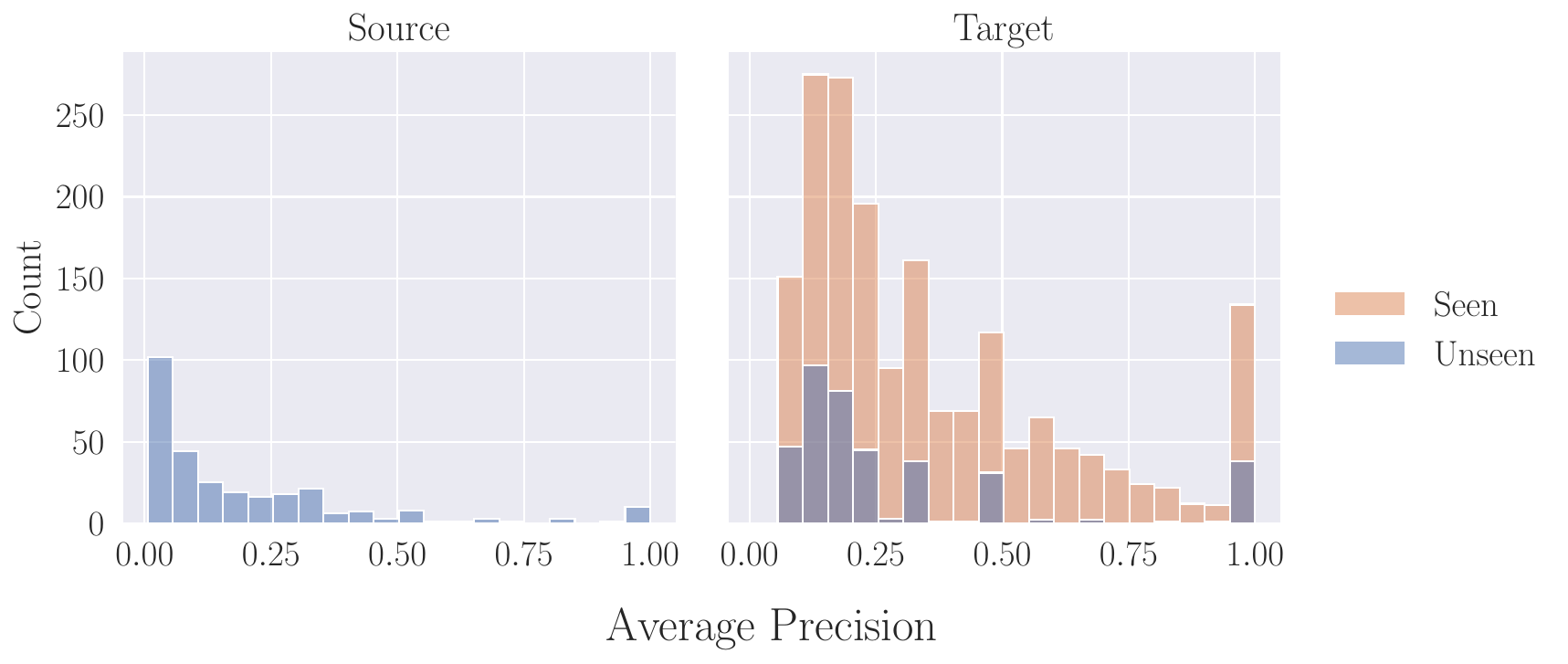}
    \caption{Average precision distributions for test set sources and targets.
    The GIN$_{CP}$ model was trained on cold-source split data. Each point in
    the histogram is a source or target node, and the color indicates whether
    that node had been seen during model training.}
    \label{fig:seen_ap}
\end{figure}

To assess the model's zero-shot prediction potential, where neither the source
nor the target nodes are seen in training or validation, we computed the average
precision (AP) scores for each source and target node to evaluate how well the
model was able to retrieve its true links. In this cold-source split scenario,
all source nodes in the test set are completely unseen during training, and most
but not all target nodes are seen (a target node may not be seen if it is only
connected to left out sources). Thus, the model's prediction of an edge with
isolated endpoints, i.e. both source and target are unseen during training, is a
zero-shot link prediction. As shown in Figure \ref{fig:seen_ap}, the presence of
unseen target nodes with high AP scores (see the rightmost blue bar in the
Target panel) shows that for some node pairs, the model is able to predict many
true links even though both the source and target nodes are isolated at test
time. This suggests the model, trained on CP features and known drug-gene
relationships, can predict links between completely novel drugs and genes.
Further work should be done to explore the efficacy of DTI predictions in the
zero-shot setting.

\section{Discussion}\label{sec:discussion}

Predicting the complex relationships between chemical compounds and genes is
ambitious. Developing models to accurately predict these high-order
interactions, particularly in inductive or zero-shot settings, is even more
challenging. The \MOTIVE{} dataset addresses these challenges by integrating
morphological features of cells and known compound-gene relationships. By
providing a large-scale, feature-rich, and extensively annotated dataset,
\MOTIVE{} enables the development and benchmarking of graph-based models for DTI
prediction in transductive, inductive, and zero-shot scenarios. One of the key
strengths of \MOTIVE{} lies in its rigor. The morphological profiles were
extracted from the JUMP CP dataset using a standardized pipeline, ensuring
consistency and reproducibility. Carefully curating annotations prioritized data
quality and reliability. The constructed dataset also offers stringent
protection against data leakage of any kind while providing thorough and
challenging forms of data splitting, loading, minibatch training (requiring
subgraphs), sampling, and evaluation procedures.

We acknowledge certain limitations in \MOTIVE{}. As discussed in Section
\ref{ablation}, the effectiveness of CP features may be reduced when the graph
is densely connected. Additionally, some perturbations may not induce
significant morphological changes, potentially leading to uninformative node
representations and false positive predictions. To address these issues, we
propose isolating samples with distinguishable morphologies and reanalyzing them
to assess the impact on performance. Finally, we recognize that the availability
of image-based profiles remains a limitation for expanding our DTI graph; we
propose that \MOTIVE{} could be extended by training a generative model that
translates the compound structure to in-silico Cell Painting readouts, as
similarly explored in Zapata et al. \citep{MarinZapata2022CellMD}. Also, if the
network is extended to include multiple modalities, then it could also be
adapted to make predictions for nodes with missing modalities
\citep{Wang2023MultiModalLW}.

Future work may incorporate both the ORF and CRISPR gene features as a form of
multimodal inputs in the graph. \MOTIVE{} could also be expanded to include
alternative representations for compound and gene nodes (e.g. protein structures
in addition to image-based profiles) to capture known structural and sequencing
relationships. The complementary information between profiles may then lead to
higher quality DTI predictions. Additionally, methods may be extended to predict
heterogeneous interactions rather than just binary classifications. Finally,
developing end-to-end architectures to learn node embeddings directly from
images could better exploit the morphological information. \MOTIVE{} represents
a valuable resource for the machine learning community, particularly for those
interested in graph-based methods and their applications in drug discovery. By
fostering interdisciplinary collaborations across graph ML, biological imaging,
and drug discovery, \MOTIVE{} has the potential to accelerate progress for the
challenging and complex task of DTI prediction.

\section*{Acknowledgments}
The authors gratefully acknowledge an internship from the Massachusetts Life
Sciences Center (to ES). We appreciate funding from the National Institutes of
Health (NIH MIRA R35 GM122547 to AEC) and AEC is a Merkin Institute Fellow at
the Broad Institute of MIT and Harvard.

\bibliographystyle{unsrtnat_short}
\bibliography{references}

\newpage
\appendix
\section*{Appendix}

\section{GraphSAGE\texorpdfstring{$_{CP}$}{\_CP} forward pass algorithm}\label{CPalg}

\begin{algorithm}[h]
\caption{G$_{CP}$ link prediction algorithm}\label{alg:gnn}
\KwIn{\begin{tabular}[t]{l}
      $G(\mathcal{V}, \mathcal{E})$ \\
      Input features $\{\mathbf{x}_s, \forall s \in \mathcal{S}\}$ and $\{\mathbf{x}_t, \forall t \in \mathcal{T}\}$ \\
      Embedding weight matrices $\mathbf{E}_s$ and $\mathbf{E}_t$ \\
      Message Passing Network $\mathbf{C}_1$ and $\mathbf{C}_2$ \\
      Non-linearity functions reLU and $\sigma$\\
      Neighborhood function $\mathcal{N} : v \rightarrow 2^{\mathcal{V}}$\\
    \end{tabular}}
\KwOut{$\mathbf{y}_{uv}, \forall (u,v) \in $ supervision edge set $ST_{sup}$}

$\mathbf{h}^0_s \gets \mathbf{x}_s\mathbf{E}_s, \forall s \in \mathcal{S}$ \Comment{Map $X_s$ to shared ft. space}\;
$\mathbf{h}^0_t \gets \mathbf{x}_t\mathbf{E}_t, \forall t \in \mathcal{T}$ \Comment{Map $X_t$ to shared ft. space}\;

\For {$v \in \mathcal{V}$}{
    $\mathbf{h}^{1}_v \gets \mathbf{C}_1(\{h_i^0 ; \forall i \in \mathcal{N}(v)\})$\;
    $\mathbf{h}^1_v \gets \mathrm{reLU}(\frac{\mathbf{h}^{1}_v}{\parallel \mathbf{h}^{1}_v\parallel_2})$\;}
\For {$v \in \mathcal{V}$}{
    $\mathbf{h}^{2}_v \gets \mathbf{C}_2(\{h_i^1 ; \forall i \in \mathcal{N}(v)\})$\;
    $\mathbf{h}^2_v \gets \frac{\mathbf{h}^{2}_v}{\parallel \mathbf{h}^{2}_v\parallel_2}$\;
    }
$\mathbf{z}_v \gets \mathbf{h}^1_v + \mathbf{h}^2_v, \forall v \in \mathcal{V}$\;

\For {$(u, v) \in ST_{sup}$}{
    $\mathbf{y}_{uv} = \sigma(\mathbf{z}_u \cdot \mathbf{z}_v )$\;
    }

\end{algorithm}

\section{Experimental details}\label{sec:experimental}

We performed a random hyperparameter search \citep{bergstra2012random} for each
model and data split for the number of hidden channels ($64, 128,256$), learning
rate ($[10^{-6}, 10^{-2}]$), and weight decay ($[10^{-5}, 1]$). We sampled
negative edges at a ratio of 1:1 at training and validation time, and 1:10 at
test time. We trained each model for 1000 epochs, computed the Binary Cross
Entropy loss between the prediction scores for the supervision edges and their
ground truth labels, and used an Adam Optimizer to make weight updates. At
validation and test time, we computed the F1 scores of the predicted edges using
the best threshold found during validation time. We also computed two rank-based
metrics, Hits@500 and Precision@500, to better capture how the model
distinguishes between positive and negative samples. Hits@$k$ quantifies the
fraction of positive test edges (total = $n$) that rank ($r$) within the top $k$
negative test edge scores
\footnote{\url{https://ogb.stanford.edu/docs/linkprop/}}. Precision@$k$
quantifies the fraction of the top $k$ predicted scores that are assigned to
true positive edges. In both cases, $k=500$ was chosen to isolate the around
$1\%$ of the test edge scores. To define these metrics more formally, let $
k^{-} $ represent the rank of the $k{\text{th}}$-ranked negative edge. Then, the
following equations apply:

\begin{align} \label{metrics}
    H_k(r_1, r_2, \dots, r_n) &= \frac{1}{n}\Sigma^n_{i=1}\mathbb{I}[r_i \leq k^{-}], \;\;
\mathbb{I} = \begin{cases}
    1 & r_i \leq k^{-}\\
    0 & r_i > k^{-}
\end{cases}
\\
P_k(r_1, r_2, \dots, r_k) &= \frac{1}{k}\Sigma^k_{i=1}\mathbb{I}[r_i], \;\;
\mathbb{I} = \begin{cases}
    1 & \text{i is a positive edge}\\
    0 & \text{i is a negative edge}
\end{cases}
\end{align}

We ran each scenario five times, and we reported the average performance for
each model as well as the standard deviation scores in Tables
\ref{tab:transductive} and \ref{tab:inductive}. We set a random seed for all
runs and established the same data split for each of our splitting methods, such
that all variance in performance must come from batch sampling by the data
loader and  GPU non-determinism. All experiments ran in a NixOS server with a
AMD Ryzen Threadripper PRO 7995WX processor and an NVIDIA RTX 6000 GPU card.

\section*{Checklist}

\begin{enumerate}

\item For all authors...
\begin{enumerate}
  \item Do the main claims made in the abstract and introduction accurately
  reflect the paper's contributions and scope?
    \answerYes{}
  \item Did you describe the limitations of your work?
    \answerYes{In Section \ref{sec:discussion}}
  \item Did you discuss any potential negative societal impacts of your work?
    \answerNo{As \MOTIVE{} relies on publicly available compound-gene
    relationships and morphological profiles and does not involve any human
    subject data, we do not believe this dataset may be misused to have a
    negative societal impact.}
  \item Have you read the ethics review guidelines and ensured that your paper
      conforms to them?
    \answerYes{}
\end{enumerate}

\item If you are including theoretical results...
\begin{enumerate}
  \item Did you state the full set of assumptions of all theoretical results?
    \answerNA{}
    \item Did you include complete proofs of all theoretical results?
    \answerNA{}
\end{enumerate}

\item If you ran experiments (e.g. for benchmarks)...
\begin{enumerate}
  \item Did you include the code, data, and instructions needed to reproduce
      the main experimental results (either in the supplemental material or as
      a URL)?
    \answerYes{In the abstract and at \url{https://github.com/carpenter-singh-lab/motive}}
  \item Did you specify all the training details (e.g., data splits,
      hyperparameters, how they were chosen)?
    \answerYes{In Appendix \ref{sec:experimental}}
    \item Did you report error bars (e.g., with respect to the random seed
        after running experiments multiple times)?
    \answerYes{We included the standard deviation scores across runs in Tables
    \ref{tab:transductive} and \ref{tab:inductive}}
    \item Did you include the total amount of compute and the type of resources
        used (e.g., type of GPUs, internal cluster, or cloud provider)?
    \answerYes{In Appendix \ref{sec:experimental}}
\end{enumerate}

\item If you are using existing assets (e.g., code, data, models) or
    curating/releasing new assets...
\begin{enumerate}
  \item If your work uses existing assets, did you cite the creators?
    \answerYes{We compiled existing databases and cited the datasets in Table
    \ref{tab:annotation_db_list}}
  \item Did you mention the license of the assets?
    \answerNo{We will discuss the licensing of dependency assets and our asset
    in the dataset documentation as part of the supplementary materials.}
  \item Did you include any new assets either in the supplemental material or
      as a URL?
    \answerYes{We will include detailed directions for how to access our dataset
    in the supplementary materials. The abstract also links to the GitHub page,
    where the asset can be accessed.}
  \item Did you discuss whether and how consent was obtained from people whose
      data you're using/curating?
    \answerYes{All of the data we used and curated was publicly available and
    published online.}
  \item Did you discuss whether the data you are using/curating contains
      personally identifiable information or offensive content?
    \answerNo{We believe we made clear from the description of our dataset that
    it does not contain personally identifiable information or offensive
    content.}
\end{enumerate}

\item If you used crowdsourcing or conducted research with human subjects...
\begin{enumerate}
  \item Did you include the full text of instructions given to participants and
      screenshots, if applicable?
    \answerNA{}
  \item Did you describe any potential participant risks, with links to
      Institutional Review Board (IRB) approvals, if applicable?
    \answerNA{}
  \item Did you include the estimated hourly wage paid to participants and the
      total amount spent on participant compensation?
    \answerNA{}
\end{enumerate}

\end{enumerate}
\end{document}